# A High-Speed Capable Spherical Robot

Bixuan Zhang[1], Fengqi Zhang[1], Haojie Chen[1], You Wang[1,4],
Jie Hao[2], Zhiyuan Luo[3], Guang Li[1]

*Abstract*—This paper designs a new spherical robot structure capable of supporting high-speed motion at up to 10 m/s. Building upon a single-pendulum-driven spherical robot, the design incorporates a momentum wheel with an axis aligned with the secondary pendulum, creating a novel spherical robot structure. Practical experiments with the physical prototype have demonstrated that this new spherical robot can achieve stable high-speed motion through simple decoupled control, which was unattainable with the original structure. The spherical robot designed for high-speed motion not only increases speed but also significantly enhances obstacle-crossing performance and terrain robustness.

## I. INTRODUCTION

Spherical robots exhibit highly flexible and efficient movement characteristics [1], with their internal mechanisms and a majority of sensors encapsulated within the protective shell, effectively isolating them from external elements such as dust and other adverse factors. The sealed construction also grants these robots an inherent amphibious capability to some extent [2][3]. Their unique driving mechanism conserves a significant amount of energy, thereby endowing them with strong endurance and extended operational runtime [4]. Due to their inherent structural attributes, spherical robots possess a natural resistance to toppling, allowing them to swiftly navigate uneven and challenging terrains, and potentially displaying a trait where they become more stable at higher speeds [5][6]. The mounting platforms on either side of the spherical robot can be customized to carry different sensors or human-machine interaction devices according to practical needs, thereby broadening their application scope and areas into sectors such as industry, agriculture, and national defense and security [7][8].

The internal configurations of spherical robots are highly diverse and intricate, as exemplified by Rollo [9] developed by Aarne Halme, GroundBot [10] created by Mattias Seeman, and the BHQ [11] and BYQ[12][13] series innovated by Sun Hanxu and co-researchers, along with subsequent iterations.

[1] The State Key Laboratory of Industrial Control Technology, Institute of Cyber Systems and Control, Zhejiang University, Hangzhou, China. (e-mail: {12432039, 12132035, 22432059, king_wy, guangli}@zju.edu.cn).

[2] The Luoteng Hangzhou Techonlogy Co.,Ltd. Hangzhou, China. (e-mail: mac@rotunbot.com).

[3] The Department of Computer Science, Royal Holloway, University of London, TW20 0EX Egham, U.K. (e-mail: Zhiyuan.Luo@cs.rhul.ac.uk).

4 Corresponding author: You Wang.

**Funding information:** This work was supported by State Key Laboratory of Industrial Control Technology. **Grant Number: ICT2024A21.**

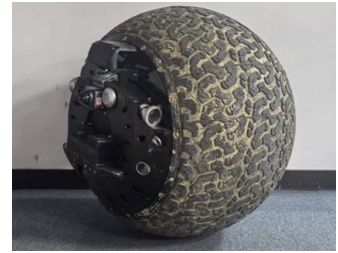

Fig. 1. Spherical robot

Most spherical robots do not incorporate redundant structures in their design, which consequently leads to strong coupling within their kinematic and dynamic models, presenting a notable characteristic of high control complexity.

Spherical robots based on a pendulum drive system stand out among various spherical robot structures due to their simple construction and compact drive components, which occupy minimal space. This design allows for greater internal volume available for expansion, offering substantial potential for applications. Consequently, a significant amount of motion control research has been carried out on this platform in recent years [14]. The spherical robot, as shown in Fig. 1, obtains propulsion through two motors driving the pendulum to swing forward, backward, left, and right. Based on this structure, a variety of algorithms have been tested.These include control using a fuzzy PID controller [15], studies using a sliding mode controller (SMC) [16][17], bias-free linear model predictive control for speed control [18], and a robust servo LQR controller with state compensation and speed feedforward for directional control [19]. The focus of these algorithms has primarily been on speed control and path following, and experiments have been conducted at speeds below 2 m/s. As speed increases further, the errors caused by linearization will increase, leading to model mismatch. Additionally, in order to maintain stability, the algorithms sacrifice some of the turning performance when increasing speed.

The structure of a momentum wheel is commonly found in the fields of aerospace and marine vessels[20]. In spherical robots, there are designs that utilize flywheels; however, these designs do not rely on the conservation of angular momentum to generate torque, but instead use the fixed-axis property of the flywheel for stabilization [21].In the domain of unmanned bicycles, there are designs that incorporate momentum wheels [22][23], but these are mainly used at low speeds. At high speeds, stabilization of the bike is primarily achieved using the handlebars.

To address the challenges of stable high-speed motion and agile steering in spherical robots,this paper innovates upon the single-pendulum-driven spherical robot by incorporating a momentum wheel aligned with the axis of rotation for the lateral swinging of the pendulum, resulting in a new structured spherical robot named RotunBot. In this new design, the main pendulum controls the rolling angle of RotunBot, while the momentum wheel regulates angular velocity of roll, thereby implementing redundant control with separated control objectives for the rolling angle direction. Experiments have demonstrated that this novel approach significantly reduces the complexity of controlling the rolling angle and simplifies the algorithm, while enhancing stability in the rolling angle direction. As a result, the maximum operating speed and turning capability are greatly improved, along with enhanced obstacle crossing ability and terrain robustness.

TABLE I
MODEL AND WEIGHT OF KEY COMPONENTS

| Component | Model | Weight(kg) |
|---|---|---|
| Heavy pendulum (with battery) | Cast steel and lead blocks | 73.4 |
| Spherical shell (with Flange plate) | Polyester fiber coated rubber | 27.6 |
| Momentum wheel | Cast steel | 9.8 |
| External components | Nylon shield and External sensors | 15.1 (both sides) |
| main shaft (with the rest components) | Aluminum alloy | 34.3 |

## II. MECHATRONIC AND CONTROL SYSTEM DESIGN

The single-pendulum-driven spherical robot is propelled by two motors that separately control the forward-backward and left-right swinging of a pendulum inside the spherical shell, causing the robot's overall center of gravity to shift and thereby enabling its motion. As a typical underactuated robot, this design utilizes the pendulum's dual-axis swinging to simultaneously manage the robot's forward-backward movement and steering. Due to the inherent instability of the spherical shape, the pendulum must not only control the robot's direction through lateral swinging but also ensure stability in the roll angle direction.

### A. Design of the hybrid wheel-pendulum structure

Building upon the single-pendulum-driven spherical robot, we innovate by incorporating a momentum wheel with an axis aligned with the axis of rotation for the lateral swinging of the pendulum, leading to the design of a new hybrid structure called RotunBot. The main advantage of RotunBot is that it separates the tasks of steering and stabilization: the pendulum no longer needs to handle both simultaneously, as the momentum wheel can take over the stabilization task, significantly reducing control complexity. The momentum wheel operates based on the principle of conservation of angular momentum, offering faster response times without altering the center of gravity, which simplifies the model and reduces control difficulty while enhancing robustness. Moreover, Compared to bicycles [22], spherical robots have more space for momentum wheels, thus allowing for the design of larger momentum wheels. Under the same requirements of rotational inertia, this results in a smaller proportion of the weight dedicated to the momentum wheels.

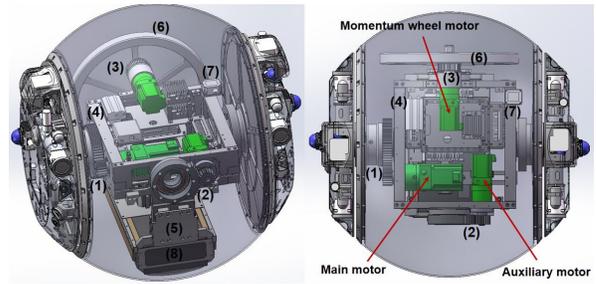

Fig. 2. Comprehensive structural diagram of RotunBot. The serial numbers correspond to specific components: (1)main drive structure, (2)auxiliary drive structure, (3)momentum wheel drive structure, (4)main shaft, (5)heavy pendulum, (6)momentum wheel, (7)Inertial Measurement Unit (IMU), (8)battery.

Specifically, whereas the momentum wheel's weight constitutes 1/3 of the total weight in [22]. In RotunBot, it accounts for less than 1/15 of the total weight while remaining effective, demonstrating the structural advantages when applied to spherical robots. RotunBot consists of main body, spherical shell, and two symmetrically arranged external components that are rigidly connected to the main body, as shown in Fig. 2. The main body of RotunBot is constructed with an aluminum alloy spindle, to which three motors are rigidly attached. Through the drive structure, these motors are connected respectively to the spherical shell, the heavy pendulum, and the momentum wheel. The battery is fixed inside the heavy pendulum, serving as part of its counterweight. Table I lists the models and weights of key components. The total mass of RotunBot $M$ is approximately 160 kilograms. The spherical shell's radius $R$ is 40 cm, the pendulum arm length $l$ is about 27 cm, and the outer diameter of the momentum wheel $d$ is 42 cm.

The composite structure of the wheel and pendulum is shown in Fig. 2. Both the main drive structure and the auxiliary drive structure utilize gear transmission mechanisms. The main drive structure (1) connects to the main motor on one end and is rigidly attached to a flange on the other, which in turn is fixed to the spherical shell. When the main motor rotates, it causes the entire assembly from (2) to (7) to rotate, altering the overall center of gravity of RotunBot, thereby enabling forward and backward movement. Notably, the external components on both sides of RotunBot are rigidly connected to the main shaft, meaning that these components will rotate with the main shaft while experiencing relative motion with respect to the flange and spherical shell.

The auxiliary drive structure (2) connects to the auxiliary motor on one end and is rigidly attached to the pendulum arm on the other. When the auxiliary motor rotates, it lifts the pendulum laterally, changing the overall center of gravity of RotunBot, causing it to tilt left or right. RotunBot completes steering tasks by moving forward and backward while in a tilted state.

The momentum wheel motor is connected to a momentum wheel through a belt drive mechanism. By controlling the acceleration and deceleration of the momentum wheel based on the principle of conservation of momentum, lateral torque is provided to the entire RotunBot system.

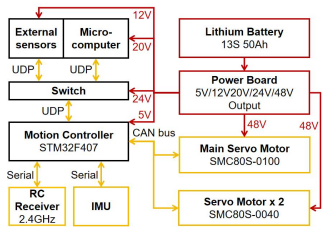

Fig. 3. Electrical architecture of RotunBot, where the yellow lines represent logic circuits, and the red lines denote power circuits.

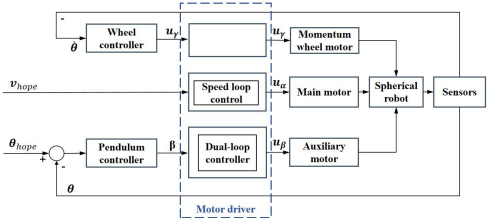

Fig. 4. Control strategy of RotunBot.

## B. Electrical architecture

Fig. 3 illustrates the electrical architecture of RotunBot. The system is powered by a 50Ah, 13S lithium battery. RotunBot supports remote control operation, using an micro-controller (STM32F407). The microcontroller connects to the remote control receiver and the IMU via serial communication and communicates with three servo motors at 100Hz through a CAN bus. Inside the RotunBot, a micro-computer can communicate with the microcontroller and other external sensors via a switch using the UDP protocol, enabling more complex tasks such as autonomous navigation, obstacle avoidance, and Simultaneous Localization and Mapping.

## C. Control strategy

Fig. 4 illustrates the control strategy for RotunBot, where $v_{hope}$ represents the desired speed, $\theta_{hope}$ is the desired roll angle, $\theta$ is the actual roll angle, $\dot{\theta}$ is the actual roll angular velocity, and $\beta$ is the angle of the heavy pendulum relative to the main shaft, with the perpendicular position set as 0. The output torques of the main motor, auxiliary motor, and momentum wheel motor are denoted by $u_\alpha$, $u_\beta$, and $u_\gamma$, respectively. To reduce model complexity while highlighting the structural advantages of RotunBot, we chose to implement independent control for each of the three motors, disregarding their mutual influences during the control process. Speed control is managed by the driver of the main motor, assuming that the motor speed is proportional to RotunBot's forward speed without considering the pitch effect of the main shaft.

Roll angle control is achieved by decoupling the model using the momentum wheel and the heavy pendulum. The heavy pendulum controls RotunBot to reach the target roll angle, while the momentum wheel stabilizes the roll angle. On this decoupled model foundation, to further highlight the structural advantages, a segmented PD controller is used for the momentum wheel, with the control objective of setting the roll angular velocity to zero. The input to this controller is the actual roll angular velocity of RotunBot, and the output is the motor torque. For the heavy pendulum, a segmented PI controller with model feedforward is employed, aiming to achieve RotunBot's target roll angle. The input to this controller is the actual roll angle of RotunBot, and the output is the angle of the heavy pendulum relative to the main shaft. This output goes through the speed and position loops of the auxiliary motor driver to generate the auxiliary motor torque. To more intuitively demonstrate the reduced complexity of controlling the heavy pendulum, the design algorithm for the feedforward controller is as follows. Define $r$ as the turning radius of RotunBot, $mg$ as the weight of the heavy pendulum, and $\beta_{pre}$ as the output of the pendulum feedforward controller. When $\theta_{hope}$ and $v$ are not zero, the gravitational moment component of the heavy pendulum provides centripetal torque, leading to the following relationship[24]:

$$mgl \sin(\beta_{pre} - \theta_{hope}) = \frac{Mv^2}{r} \quad (2)$$

$$r = \frac{R}{\tan \theta_{hope}}$$

By solving the two equations simultaneously, we obtain the results for the model feedforward calculation:

$$\beta_{pre} = \theta_{hope} + \sin^{-1}\left(\frac{Mv^2}{mgl} \tan(\theta_{hope})\right) \quad (3)$$

This control strategy employs basic feedback control to the greatest extent possible. By simplifying the control approach, it highlights RotunBot's significant advantages in reducing control complexity, enhancing system stability, and particularly its stability during high-speed motion.

## III. EXPERIMENTAL VALIDATION

Fig. 1 showcases the platform used for conducting real-world experiments.

### A. Stability comparison

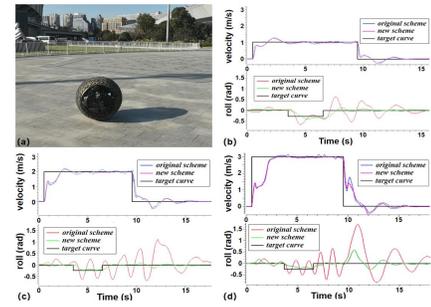

Fig. 5. (a) Real-world control performance comparison experiment. (b), (c), and (d) present comparisons of the velocity v and roll angle θ changes between the original and new control schemes at speeds of 1 m/s, 2 m/s, and 3 m/s, respectively.

The experimental site for the stability comparison experiment is shown in Fig. 5(a). To control variables, we compare the control schemes with and without the momentum wheel on the same RotunBot platform, which has identical weight and hardware configurations. Due to space limitations, step commands for target speeds of 1 m/s, 2 m/s, and 3 m/s are issued to make it move in a straight line for 9 seconds each. During the period from 3s to 6s, a step command for a target roll angle of 15 degrees (0.26 rad) is issued, observing the roll angle stability of RotunBot during acceleration, deceleration, and continuous turning.

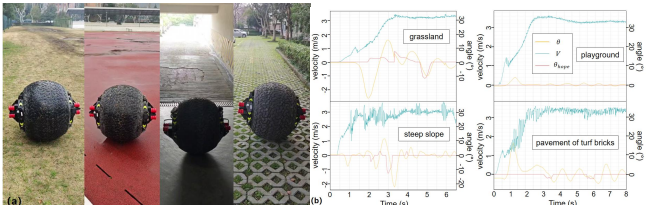

Fig. 6. (a) Real-world Terrain robustness experiment in grassland, playground, 10° steep slope, and pavement of turf bricks. (b) The changes in velocity v, hope roll angle $\theta_{hope}$ and roll angle $\theta$ during the test shown in (a).

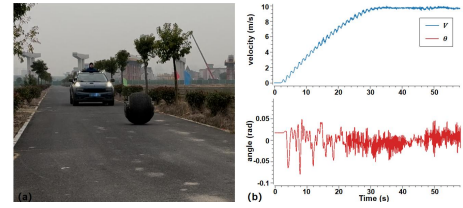

Fig. 8. (a) Real-world test for maximum stable speed. (b) The changes in velocity v and roll angle $\theta$ during the test shown in (a).

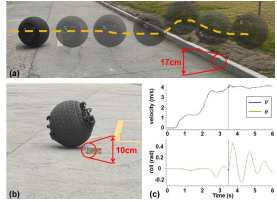

Fig. 7. (a) Real-world obstacle-crossing experiment where RotunBot crosses a 17 cm high step. (b) Real-world disturbance resistance experiment where RotunBot is subjected to lateral disturbance from a 10 cm high brick while moving at high speed. (c) The changes in velocity v and roll angle $\theta$ during the test shown in (b), with the blue vertical line indicating the moment when RotunBot collides with the obstacle marked by the red circle in (b).

Fig. 5(b) through 5(d) show the velocity and roll angle data for RotunBot at speeds of 1 m/s, 2 m/s, and 3 m/s, respectively. The blue and red curves represent the velocity and roll angle of the original control scheme without momentum wheel, while the purple and green curves represent the velocity and roll angle of the new control scheme with momentum wheel. Comparing the roll angle curves in (b)-(d), the new control scheme at 1 m/s speed shows an increase in response time by approximately 0.3 seconds, but the overshoot decreases by over 70%. At speeds of 2 m/s and above, it successfully converges, maintaining stability during continuous turning tasks without significant oscillation, and the speed curve appears smoother without glitches due to stable movement. As seen in (d), when the speed increases, RotunBot still experiences divergence in the roll angle during deceleration, but the peak value decreases by 60%, and the convergence time is reduced by about 5 seconds. The experiment indicates that the new scheme with the addition of momentum wheels has significantly improved stability of the roll angle, and the effect of controlling roll angle stability does not weaken as speed increases.

*B. Terrain Robustness*

We test the terrain robustness of RotunBot on four different surfaces, as shown in Fig. 6(a), with results presented in Fig. 6(b). In these results, the green curve represents the velocity $v$, the red curve indicates the target roll angle $\theta_{hope}$, and the yellow curve shows the actual roll angle $\theta$. Due to safety considerations and terrain limitations, we conduct the experiment by running RotunBot straight at a target speed of 3.5 m/s for approximately 10 seconds to evaluate its terrain robustness. On the three terrains outside the plastic track, where the areas are narrow, we make direction adjustments via remote control. The experiments confirm that RotunBot can operate stably across various terrains without experiencing roll angle divergence, demonstrating robust performance on different grounds. Additionally, a RotunBot modified external components shows good adaptability in amphibious environments.

*C. Obstacle Crossing and Disturbance Resistance*

We test RotunBot's obstacle-crossing ability using a 17 cm high step between a road and grass, as shown in Fig. 7(a). RotunBot successfully crosses the step by accelerating to gain momentum, briefly becoming airborne before landing and quickly resuming stable motion. The disturbance resistance test, depicted in Fig. 7(b), involves RotunBot colliding with a 10 cm high stack of bricks at 4 m/s. As seen in Fig. 7(c), although the collision causes a roll angle peak exceeding 0.4 rad, the oscillation reduces below 0.1 rad within 2 seconds. These tests show that RotunBot can operate at higher speeds while maintaining good obstacle-crossing and disturbance resistance capabilities, significantly broadening its potential applications and highlighting its robust design advantages.

*D. Maximum Stable Speed*

To understand RotunBot's maximum stable speed, we upgrade the main motor gearbox to increase its theoretical maximum speed while retaining acceleration capability and test it on a straight road, as shown in Fig. 8(a). Due to the high speed, we followed in a vehicle to ensure that RotunBot accelerated in a straight line to its maximum speed and maintained this speed for over 20 seconds. As shown in Fig. 8(b), RotunBot's maximum speed stabilizes at approximately 10 m/s, with the roll angle remaining very stable during both acceleration and steady-speed phases. Although vibration frequency increases at top speed, it remains within 0.05 rad. Furthermore, RotunBot responds to small directional corrections, enabling it to travel in a straight line.

## IV. CONCLUSION

This paper introduces a novel spherical robot, RotunBot, designed for high-speed stable operation and turning capabilities. By integrating a momentum wheel with the pendulum-driven mechanism, RotunBot achieves separate control of turning and roll angle stability, significantly reducing control complexity and enhancing stability and maximum speed. The study covers the design of its electrical architecture and control strategy, with experiments confirming improved motion stability, disturbance resistance, and effective turns at higher speeds. Enhanced stability and speed also boost RotunBot's hill-climbing, obstacle-crossing performance, and terrain robustness, making it a more adaptable platform for complex environments. Future work will address instability during rapid deceleration and downhill motion, develop smarter autonomous algorithms, and explore broader application scenarios.